\documentclass{article}

\usepackage{microtype}
\usepackage{graphicx}
\usepackage{subfigure}
\usepackage{booktabs} 
\usepackage{epsfig}
\usepackage{amsmath}
\usepackage{amssymb}
\usepackage{tabu}
\usepackage{makecell}
\usepackage{color}
\usepackage{array}
\usepackage{xcolor}
\usepackage{hyperref}

\usepackage[accepted]{icml2020}

\icmltitlerunning{Latent Space Factorisation and Manipulation via Matrix Subspace Projection}

\begin{document}

\twocolumn[
\icmltitle{Latent Space Factorisation and Manipulation via Matrix Subspace Projection
}



\icmlsetsymbol{equal}{*}

\begin{icmlauthorlist}
\icmlauthor{Xiao Li}{to,goo}
\icmlauthor{Chenghua Lin}{goo}
\icmlauthor{Ruizhe Li}{goo}
\icmlauthor{Chaozheng Wang}{to}
\icmlauthor{Frank Guerin}{ed}
\end{icmlauthorlist}

\icmlaffiliation{to}{Department of Computing Science, University of Aberdeen, UK}
\icmlaffiliation{goo}{Department of Computer Science, University of Sheffield, UK}
\icmlaffiliation{ed}{Department of Computer Science, University of Surrey, UK}

\icmlcorrespondingauthor{Chenghua Lin}{c.lin@sheffield.ac.uk}

\icmlkeywords{MSP, generative models, synthesis, disentanglement, facial attribute editing,
attribute style manipulation}

\vskip 0.3in
]



\printAffiliationsAndNotice{}  

\begin{abstract}
We tackle the problem disentangling the latent space of an autoencoder in order to separate labelled attribute information from other characteristic information. This then allows us to change selected attributes while preserving other information. Our method, matrix subspace projection, is much simpler than previous approaches to latent space factorisation, for  example not requiring multiple discriminators or a careful weighting among their loss functions.
Furthermore our new model can be applied to autoencoders as a plugin, and works across diverse domains such as images or text. We demonstrate the utility of our method for attribute manipulation in autoencoders trained across varied domains, using both human evaluation and automated methods. The quality of generation of our new model (e.g. reconstruction, conditional generation) is highly competitive to a number of strong baselines. 

\end{abstract}

\section{Introduction}

We investigate the problem of manipulating multiple attributes of  data samples. This can be applied to image data, for example to manipulate a picture of a face to add a beard, change gender, or age. It can also be applied to text, for example to change the style or sentiment of a text. We assume that we have a training dataset where attributes are labelled. However there is an unsupervised aspect because we do not have samples of the same individual with different attribute combinations, e.g., the same person with and without a beard. Furthermore the training samples have some attribute combinations that are highly correlated, while other combinations are completely absent; e.g., in the CelebA dataset blond hair and earrings are highly correlated with female \cite{celebacorrel}, while a female with beard is absent. Nevertheless we would like our system to somehow isolate the explanatory factors in pixel space, to understand, e.g., that blond hair corresponds only to colour changes to hair pixels, and no change elsewhere in the face.

This challenge of isolating multiple explanatory factors poses interesting problems for generative models.
In images of faces for example, even if the training data has no bearded lady, a good generative model should be able to `imagine' novel examples that combine attributes in ways not present in training data.
As noted by \citet{HigginsBVAEearly} ``Models are unable to generalise to data outside of the convex hull of the training
distribution \ldots unless they learn about the data generative factors and recombine them
in novel ways.'' 
Ideally we should fully disentangle and isolate the data generative factors, so that we can represent the generative factors of a sample with a vector that has one part labelled attribute information, and another part with the other characteristic information of the sample. 
This is in a small way part of a general trend  to try to move deep neural network research towards explanatory models of the world \cite{LeCunpower2018,LakeBBSarxiv2016,DBLP:journals/corr/abs-1805-04025}, which requires disentanglement.
The problem is important because isolating explanatory factors is a way to overcome the combinatorial explosion  of required training examples if such factors are not isolated \cite{DBLP:journals/corr/abs-1805-04025}.

\begin{figure}[tb]
    \centering
    \includegraphics[height=2.8cm]{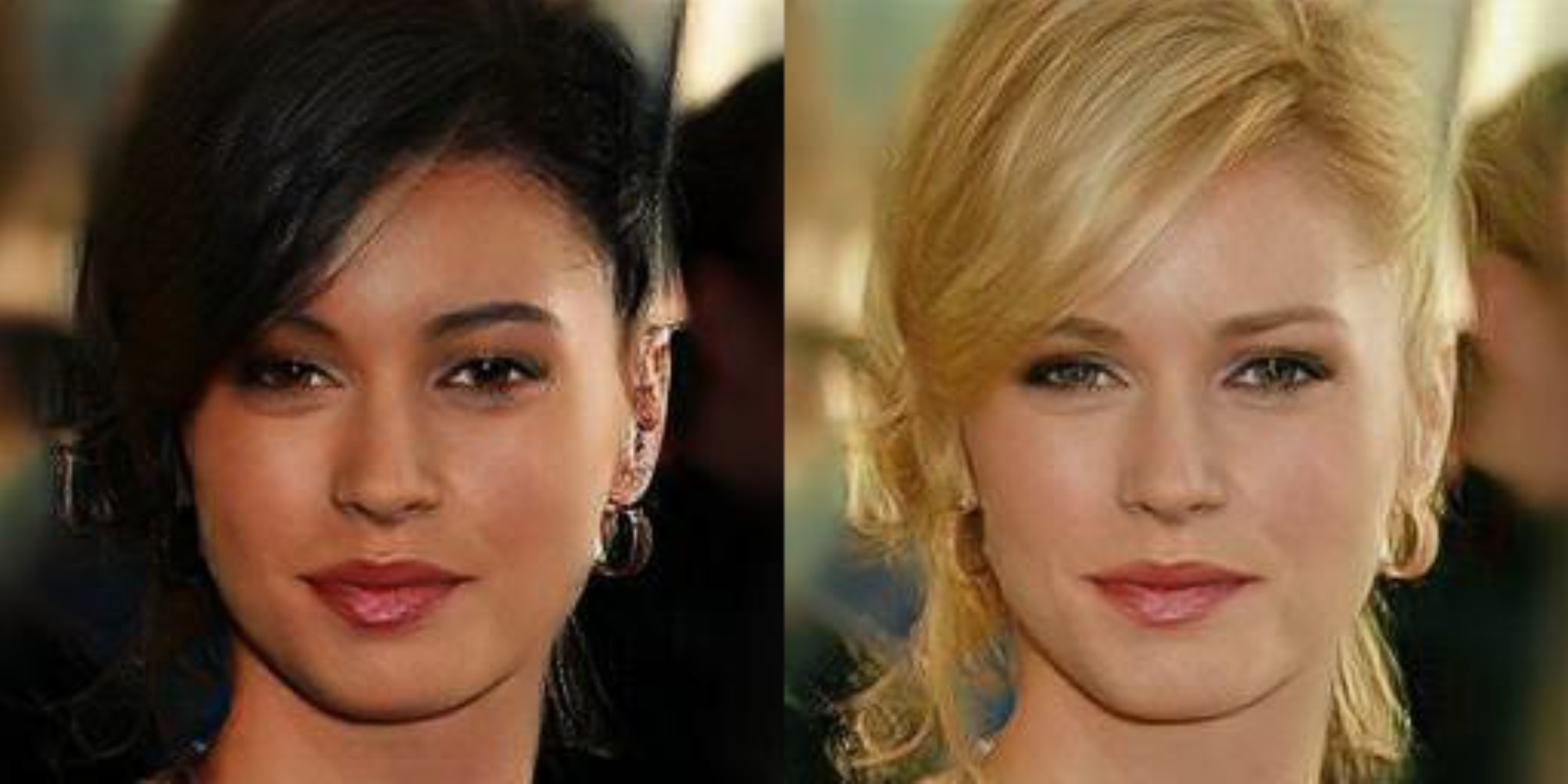}
    \includegraphics[height=2.8cm]{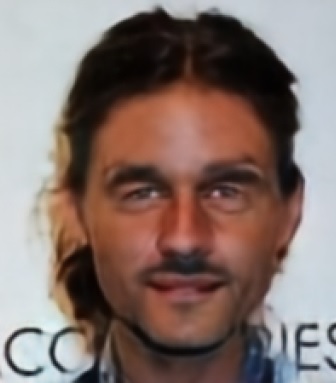}
    \caption{Left: from RelGAN \cite{relGAN}, where the only attribute changed is hair colour, but we see significant changes in skin colour, eyebrows, eyes, and lips. Right: from Fader \cite{NIPS2017_7178}, where female was changed to male, but female eyebrows are retained above the male ones, due to skip connections.}
    \label{relgan}
\end{figure}

A typical approach to the problem uses an autoencoder (AE) which encodes a given input (e.g. picture, text, etc.) into a  \textit{latent vector}, and then restores (decodes) the latent vector to the given input \cite{NIPS2017_7178,pmlr-v70-hu17e,DBLP:journals/corr/abs-1811-00135,li2019stable}. The latent vector contains the attribute information as well as other characteristic information of the  input. If one can change the attribute information in the latent space, then one can generate examples with the altered attributes. The difficulty here is twofold: (1) learn a latent space representation which separates the attributes from all other characteristic information, and (2) fully disentangle the attributes. If we fail in the separation part, then efforts to generate with specific attributes may conflict with other information in the latent space (as in \citet{NIPS2014_5352} etc., see \S\ref{sec:relatedwork}). If we fail in the second part then  examples generated with specified attributes will also be contaminated with spurious attributes (see Fig.~\ref{relgan} Left).

Many recent approaches make use of auxiliary neural network structures with adversarial training in the style of
Generative Adversarial Networks (GANs). These new networks can be used to remove attribute information from the latent space \cite{NIPS2017_7178}, or to feedback a loss term to impose the attributes they want to appear in the output \cite{attgan}.
These adversarial approaches have competing loss terms (for example reconstruction loss, attribute classification loss, realistic output loss), and require a careful choice of hyperparameters to weight these loss functions. In the case of \citet{NIPS2017_7178}  a slowly increasing loss was critical. These hyperparameters and training schedules must be determined by trial and error, to avoid training instability. Even after successful training we have found that some models ignore the desired attributes and put too much weight on reconstruction and realistic output (see \S\ref{eval}). This is partly because we push systems to the very difficult setting of training for multiple attributes together (e.g. 40 attributes for CelebA). This is a very demanding setting for disentanglement, e.g. to dissociate lipstick, make-up, and blond hair from female, and to dissociate beard, bushy eyebrows, and 5 o'clock shadow
from male.

We propose a simple and generic method, Matrix Subspace Projection (MSP), which directly separates the attribute information from all other non-attribute information, without relying on weighting loss terms from auxiliary neural networks. 
Our variables representing attributes  are fully disentangled, with one isolated variable for each attribute of the training set.
Therefore, when we do  conditional generation, we can assign  pure attributes combined with other latent data which does not conflict, so that the generated pictures are of high quality and not
contaminated with spurious attributes.
Meanwhile, our model is a universal plugin. In theory, it can be applied to any existing AEs (if and only if the AEs use a latent vector). If the AE is a generative model (such as VAE), with our approach, it becomes a conditional generative model that generates content based on the given condition constraints.
In the case of images, we add a PatchGAN at the end of our generator to sharpen the image, but this is not connected with the attribute manipulation task and is not core to our model; it could be replaced with any super resolution and sharpening method.

Our plugin has two uses: (1)   samples can be  generated from a random seed, but with given attributes; (2) a given sample can be modified to have desired  specified attributes. Our key contributions are:
(1)~A simple and universal plugin for conditional generation and content replacement, directly applicable to any AE architectures (e.g., image or text). (2)~Strong performance on learning disentangled latent representations of multiple (e.g. 40) attributes. (3)~A principled weighting strategy for combining loss terms for training. The code for our model is available online\footnote{Code: \url{https://xiao.ac/proj/msp}}.

\section{Related Work}\label{sec:relatedwork}

The first approaches to control of generation by attributes (conditional VAEs \cite{NIPS2014_5352,NIPS2015_5775,DBLP:conf/eccv/YanYSL16}) simply added attribute  information as an extra input to the encoder or the decoder. These approaches generate using a latent vector $\mathbf{z}$ and also an attribute vector $\mathbf{y}$, where the $\mathbf{z}$ often conflicts with $\mathbf{y}$, because attribute information has not been removed from $\mathbf{z}$.
With conflicting inputs the best the VAE can do is to produce a blurry image. 

Generative Adversarial Networks (GANs) can be augmented with encoders. IcGAN  trains separate encoders for the $\mathbf{y}$ and $\mathbf{z}$ vectors, but does not try to remove potentially conflicting information \cite{DBLP:journals/corr/PerarnauWRA16}. The IcGAN authors also note that it can fail to generate unusual attribute combinations such as a woman with a moustache, because the GAN discriminator  discourages the generator from generating samples outside the training distribution.
  
\begin{figure}[htb]
    \centering
    \includegraphics[width=8cm]{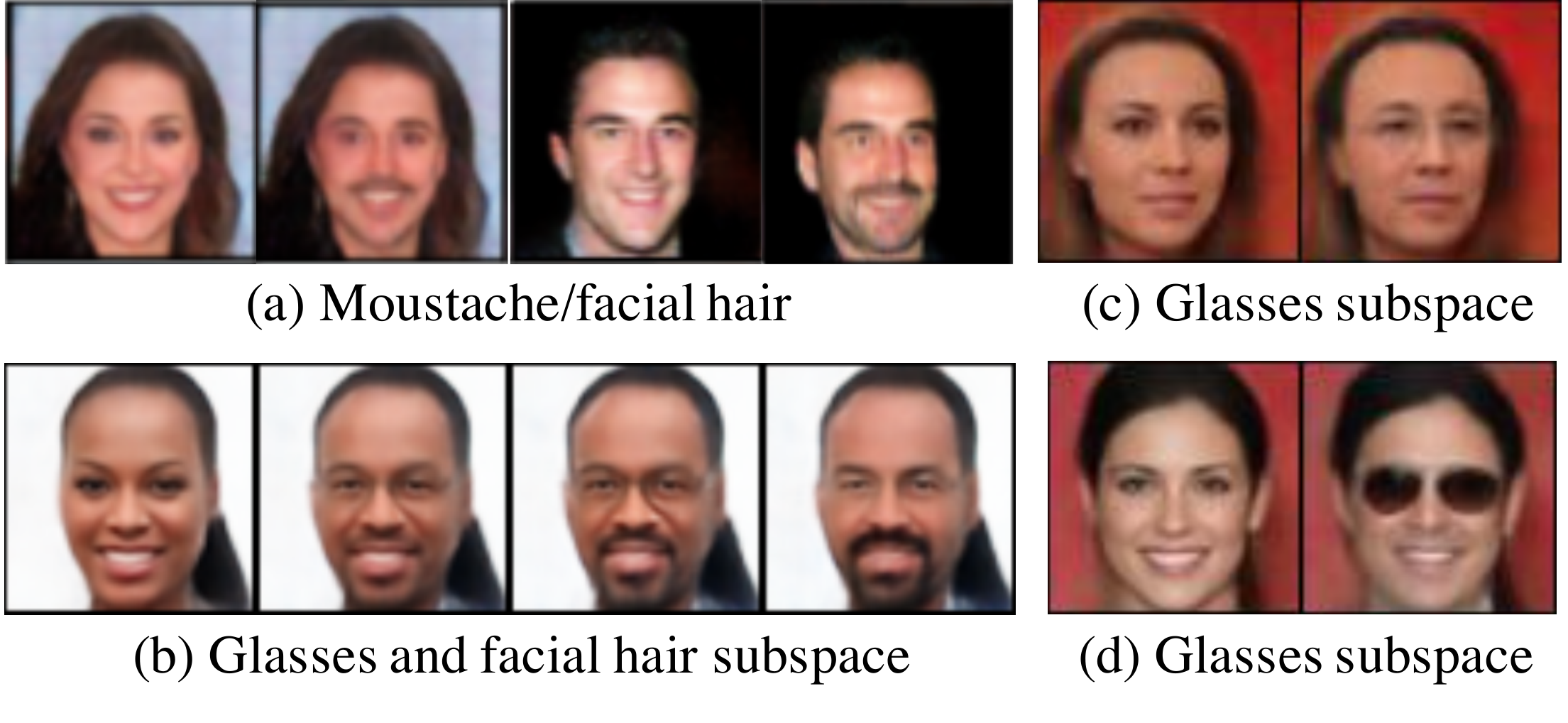}
    \caption{Figure showing the  difficulty of disentangling attributes for  supervised `adversarial' approaches.
    (a) from \citet{CreswellCondVAE-info} shows significant change in the eyebrows and eyes when adding facial hair. (b,c,d) from \citet{Klys2018NIPS}. (b) moving across the glasses and facial hair subspace, from the female on the left, brings significant changes in eyebrows and eyes, and the shape of cheeks, making the face more masculine. (c) moving in glasses subspace shows changes around the eyes and mouth, looking older. (d) also moving in glasses subspace shows a narrower smile and a more masculine lower face.
    Note all of the pictures are generated by VAE, none are original photographs.}
    \label{otherim}
\end{figure}  
  
More recent work tackled the problem of separating the attribute information from the latent vector, using a new auxiliary network  (like a GAN  discriminator)  \cite{NIPS2017_7178,CreswellCondVAE-info,Klys2018NIPS},  which attempts to guess the attribute of the latent vector $\mathbf{z}$, and penalise the generator if attribute information remains. A significant drawback of these  adversarial approaches is that great care must be taken in training so that the loss from the discriminator (which is trying to remove attribute information) does not disturb the training to produce a good reconstruction. In the case of 
Fader networks~\cite{NIPS2017_7178} it was necessary to start with a discriminator loss weight of zero, and linearly increase to 0.0001 over the first 500,000 iterations; the authors state ``\textit{This scheduling turned out to be critical in our experiments.
Without it, we observed that the encoder was too affected by the loss coming from the discriminator, even for low values of [loss coefficient]}.''

While this adversarial approach can successfully remove attribute information from $\mathbf{z}$, there is nothing to stop the decoder (generator) from associating other spurious information with the attribute. For example the decoder might associate the attribute intended to be for `glasses' with an older or more masculine face. This is what we see in the results of two of the adversarial approaches (see Fig.~\ref{otherim}). 
Most of the results in \citet{CreswellCondVAE-info} focus on the attribute `smiling' (not reproduced here), and this is very well disentangled. It is only when the training dataset associates other attributes with the trained attribute that entanglement will arise.
In \citet{CreswellCondVAE-info} the attribute vector is a single binary variable so that the system can only be trained to control (or classify) one attribute. 
It is not unexpected that a generator will associate spurious information with an attribute if the association is present in the training data and the system has  been trained only on examples labelling a single attribute, e.g., glasses. 
 The system cannot know that it should isolate `wearing glasses', and not `wearing glasses and older'.
Fader Networks  \cite{NIPS2017_7178}  can train for multiple attributes together, however
\citet{attgan} state that ``Although
Fader Networks is capable for multiple attribute editing
with one model, in practice, multiple attribute setting makes
the results blurry.''

The most recent works (2018-19) are GAN-based. They do not try to remove attribute information from the latent space, but instead add an additional attribute classifier after generation, and impose an attribute classification loss. This is in addition to a typical GAN discriminator for realistic images.
AttGAN  \cite{attgan} uses an endoder, decoder (generator), and the attribute classifier and discriminator applied to the output of the generator.
StarGAN  \cite{StarGAN} and RelGAN \cite{relGAN} use no encoder, but use a singe generator twice, in a cycle; the first direction alters attributes like a conditional GAN, the second one attempts to reconstruct the image (using original attributes), and so requires that non-attribute information has been preserved. StarGAN uses a  discriminator and attribute classifier, like AttGAN, while 
RelGAN adds a third network for interpolation.

All the works cited from 2017 to 2019 have an  adversarial component (in the style of a GAN); they train auxiliary classifiers to feed back loss terms, to ensure they remove undesirable attributes, or enforce desired ones. They need a careful weighting among loss terms, but there is no principled method for determining these weighting hyperparameters. Our work does not rely on an adversarial component to manipulate attributes; we use a more direct method of matrix projection onto subspaces, in order to factorise the latent representation and separate attributes from other information. 
Furthermore, unlike the above works\footnote{Not mentioned in the Fader networks paper, but in the published code: https://github.com/facebookresearch/FaderNetworks}
 we do not use any skip connections. Skip connections can introduce errors when a region of the source and target image is quite different, we illustrate this further in Fig.~\ref{relgan} Right.

In addition to the above works using labelled attributes there is also work on the more difficult problem of unsupervised learning of disentangled generative factors of data
\cite{NIPS2016_6399,DBLP:conf/iclr/HigginsMPBGBML17,DBLP:conf/iclr/0001SB18}.
However the supervised (labelled) approaches generate much clearer samples of selected attributes, and superior disentanglement. 
An alternative approach to controlled generation is to simply train a deep convolutional network and do linear interpolation in deep feature space \cite{DBLP:conf/cvpr/UpchurchGPPSBW17}. This shows surprisingly good results, but in changing an attribute that should only affect a local area it can affect more image regions, and can produce unrealistic results for more rare face poses.

\section{Method}

\subsection{Problem Formulation} \label{sec:problem}
We are interested in factorising and manipulating multiple attributes from a latent representation learned by an \textit{arbitrary} Autoencoder (AE). Suppose we are given a dataset $\mathcal{D}$ of elements $(\mathbf{x}, \mathbf{y})$ with $\mathbf{x} \in \mathbb{R}^{n}$ and $\mathbf{y} \in Y = \{ 0, 1\}^{k}$ representing $k$ attributes of $\mathbf{x}$. 

\begin{figure}[tb]
\hspace*{-0.4cm}
    \centering
    \includegraphics[width=8.7cm,trim={0 0.9cm 10 0.9cm},clip]{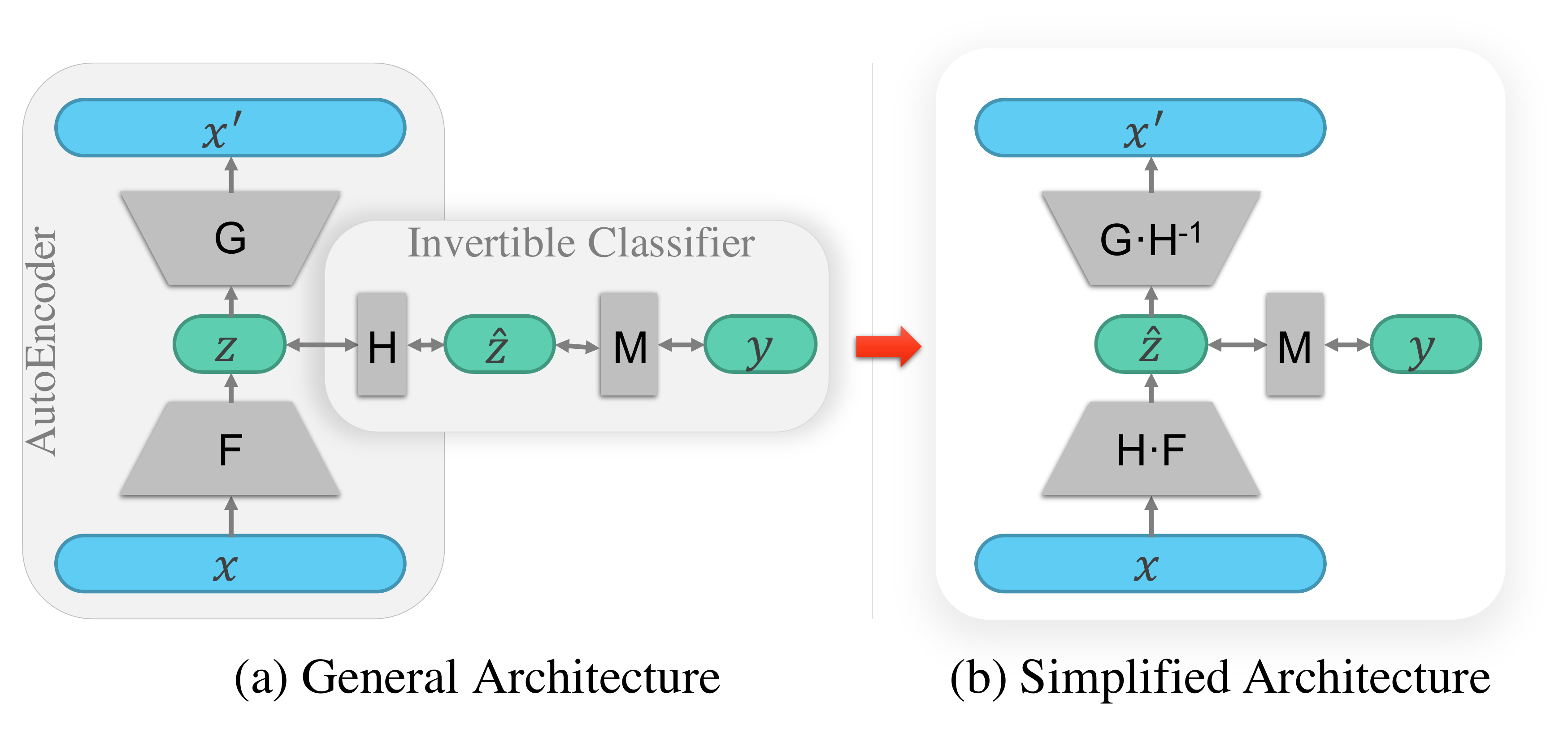}\vspace{-2ex}
    \caption{(a) The general architecture of our model (MSP); (b) the simplified architecture of MSP.}
    \label{fig:model}
\end{figure}

Let an arbitrary AE be represented by $\mathbf{z}=F(\mathbf{x})$ and $\mathbf{x'}=G(\mathbf{z})$, where $F(\cdot)$ is the encoder, $G(\cdot)$ is the decoder, $\mathbf{z}$ is the latent vector encoding $\mathbf{x}$, and $\mathbf{x'}$ is the reconstruction of $\mathbf{x}$ (see Fig.~\ref{fig:model}). Note that when $\mathbf{x'}$ is a good approximation of $\mathbf{x}$ (i.e., $\mathbf{x'}\approx \mathbf{x}$), the attribute information of $\mathbf{x}$ represented in $\mathbf{y}$ will also be captured in the latent encoding $\mathbf{z}$. 
Attribute manipulation means that we 
replace the attributes $\mathbf{y}$ captured by $\mathbf{z}$ with new attributes $\mathbf{y_n}$. Let $K(\cdot)$ be a replacement function, then we have new latent space $\mathbf{z_n} = K(\mathbf{z}, \mathbf{y_n})$ and $\mathbf{x_n}=G(K(\mathbf{z}, \mathbf{y_n}))$, where the attribute information encoded in $\mathbf{y_n}$ can be predicted from $\mathbf{x_n}$ and the non-attribute information of  $\mathbf{x}$ will be \textit{preserved}. To give a concrete example, given an image of a face, $\mathbf{x}$, we wish to manipulate $\mathbf{x}$ w.r.t. the presence or absence of a set of desired attributes encoded in $\mathbf{y_n}$ (e.g., a face with or without smiles, wearing or not wearing glasses), producing the manipulated image $\mathbf{x_n}$, without changing the identify of the face (i.e., preserving the non-attribute information of $\mathbf{x}$).

\subsection{Learning Disentangled Latent Representations via Matrix Subspace Projection}

To tackle the problem formulated in $\S\ref{sec:problem}$, we propose a generic method to factor out the information about attributes $\mathbf{y}$ from $\mathbf{z}$ based on the idea of performing orthogonal matrix projection onto subspaces.  Our model works as a universal plugin and in theory, it can be applied to any existing AEs. The general architecture of the proposed MSP model is depicted in Fig.~\ref{fig:model} (a). 
Given a latent vector $\mathbf{z}$ encoding $\mathbf{x}$ and an arbitrarily complex invertible function $H(\cdot)$, $H(\cdot)$ transforms $\mathbf{z}$ to a new linear space ($\mathbf{\hat{z}}=H(\mathbf{z})$) such that one can find a matrix $\mathbf{M}$ where  
(a) the projection of $\mathbf{\hat{z}}$ on $\mathbf{M}$ (denoted by $\mathbf{\hat{y}}$) approaches  $\mathbf{y}$ (i.e., $\mathbf{\hat{y}}$ captures attribute information), 
\begin{equation}
   \mathbf{M}\cdot \mathbf{\hat{z}}=\mathbf{\hat{y}};~~~\mathbf{\hat{y}} \rightarrow \mathbf{y}
   \label{eq:def1} 
\end{equation}
and (b) there is an \textit{orthogonal matrix} $\mathbf{U} \equiv  [\mathbf{M};\mathbf{N}]$, where $\mathbf{N}$ is the \textit{null space} of $\mathbf{M}$ (i.e., $\mathbf{M} \perp \mathbf{N}$) and the projection of $\mathbf{\hat{z}}$ on $\mathbf{N}$ (denoted by $\mathbf{\hat{s}}$) captures non-attribute information. 
As $\mathbf{U}$ is orthogonal, we also have $\mathbf{U}^T \equiv \mathbf{U}^{-1}$.

Fig.~\ref{fig:model} (b) presents a simplified architecture of our MSP model, which is equivalent to the general architecture. This simplification exists because as explained earlier $H(\cdot)$ is invertible. So  when the encoder and decoder of an AE have enough capacity, they can essentially absorb $H$ and $H^{-1}$. In other words, rather than fitting $F$ and $G$, the encoder and decoder will fit $H\cdot F(\cdot)$ and $G\cdot H^{-1}(\cdot)$ instead. As $\mathbf{M}$ itself is an incomplete orthogonal matrix, similar operations cannot be applied to $\mathbf{M}$.

Our main learning objective, in addition to the original AE objective (i.e. reconstruction loss $\mathcal{L}_{AE}$; see \S\ref{sec:totalLoss} and Eq.~\ref{eq:totalLoss}), is then to estimate $\mathbf{M}$ which is nontrivial. We turn the problem of finding an optimal solution for $\mathbf{M}$ into an optimisation problem, in which we need to (1) enforce $\mathbf{\hat{y}}$ to be as close to $\mathbf{y}$ (i.e., the vector encoding the ground truth attributes) as possible; and (2) minimise $||\mathbf{\hat{s}}||^2$  so that $\mathbf{\hat{s}}$ contains as little information from $\mathbf{\hat{z}}$ as possible. This can be formulated into the following loss function 
\begin{align}
    \mathcal{L}_\mathit{MSP} & =  \mathcal{L}_1+\mathcal{L}_2 \\
    \mathcal{L}_1 & = ||\mathbf{\hat{y}}-\mathbf{y}||^2 = ||\mathbf{M}\cdot\mathbf{\hat{z}}-\mathbf{y}||^2 \\
    \mathcal{L}_2 & = ||\mathbf{\hat{s}}||^2
\end{align}
Here $\mathcal{L}_1$ and $\mathcal{L}_2$ encode the above two constraints, respectively, and $\mathbf{\hat{y}}$ is the predicted attributes. 
Given that the AE relies on the information of $\mathbf{\hat{z}}$ to reconstruct $\mathbf{x}$, the optimisation constraints of $\mathcal{L}_{AE}$ and $\mathcal{L}_2$ essentially introduce an adversarial process: on the one hand, it discourages any information of $\mathbf{\hat{z}}$ to be stored in $\mathbf{\hat{s}}$ due to the penalty from $\mathcal{L}_2$; on the other hand, the AE requires information from $\mathbf{\hat{z}}$ to reconstruct $\mathbf{x}$.
So, the best solution is to only restore the essential information for reconstruction (except the attribute information) in $\mathbf{\hat{s}}$. 
By optimising $\mathcal{L}_\mathit{MSP}$, we cause $\mathbf{\hat{z}}$ to be factorised, with the attribute information stored in $\mathbf{\hat{y}}$, while $\mathbf{\hat{s}}$ only retains non-attribute information.

The first part of our loss function $\mathcal{L}_1$ is relatively straightforward. The main obstacle here is to compute $\mathcal{L}_2$ as $\mathbf{\hat{s}}$ is unknown. We develop a strategy to compute $||\mathbf{\hat{s}}||^2$ indirectly. According to the definition of $\mathbf{\hat{y}}$ and $\mathbf{\hat{s}}$ we can derive:
\begin{align}\label{eq:S_6-7}
     \mathcal{L}_2 = & ||\mathbf{\hat{s}}||^2=||\mathbf{\hat{s}} - \mathbf{0}||^2 \nonumber\\
    = & ||[\mathbf{\hat{y}};\mathbf{\hat{s}}]-[\mathbf{\hat{y}};\mathbf{0}]||^2~~~\text{Identical deformation} \nonumber \\ 
    = & ||\mathbf{U}\cdot \mathbf{\hat{z}}-[\mathbf{\hat{y}};\mathbf{0}]||^2
\end{align}
where the square brackets represent the vector concatenation.
Because $\mathbf{U}$ is orthogonal, we have
\begin{align}\label{eq:S_2}
     \mathcal{L}_2 =& ||\mathbf{U}\cdot \mathbf{\hat{z}}-[\mathbf{\hat{y}};\mathbf{0}]||^2 \nonumber\\
    = & ||\mathbf{U}^{-1} \cdot (\mathbf{U}\cdot \mathbf{\hat{z}}-[\mathbf{\hat{y}};\mathbf{0}])||^2 \nonumber\\ 
   = & ||\mathbf{\hat{z}}-\mathbf{U}^{-1}\cdot[\mathbf{\hat{y}};\mathbf{0}]||^2 
   = ||\mathbf{\hat{z}}-\mathbf{U}^{T}\cdot[\mathbf{\hat{y}};\mathbf{0}]||^2 \nonumber\\
   = & ||\mathbf{\hat{z}}-[\mathbf{M};\mathbf{N}]^{T}\cdot[\mathbf{\hat{y}};\mathbf{0}]||^2 \nonumber\\
   = & ||\mathbf{\hat{z}}-\mathbf{M}^{T}\cdot \mathbf{\hat{y}}||^2 
   \approx ||\mathbf{\hat{z}}-\mathbf{M}^{T}\cdot \mathbf{y}||^2
\end{align}
With Eq.~\ref{eq:S_2} (which makes use of the properties of orthogonal matrices), we avoid computing $\mathbf{\hat{s}}$ and $\mathbf{N}$ directly when minimising $||\mathbf{\hat{s}}||^2$, and turn the minimisation problem into optimising $\mathbf{M}$ instead. 
Finally, we have:
\begin{equation}\label{eq:final_l1_l2}
\begin{aligned}
   \mathcal{L}_\mathit{MSP}=& ~\mathcal{L}_1+\mathcal{L}_2 \\
   =& ~||\mathbf{M}\cdot \mathbf{\hat{z}} - \mathbf{y}||^2 + ||\mathbf{\hat{z}}-\mathbf{M}^{T}\cdot \mathbf{\hat{y}}||^2  \\ 
   \approx & ~||\mathbf{M}\cdot \mathbf{\hat{z}} - \mathbf{y}||^2 + ||\mathbf{\hat{z}}-\mathbf{M}^{T}\cdot \mathbf{y}||^2 
\end{aligned}
\end{equation}

The loss function in Eq.~\ref{eq:final_l1_l2} also guarantees that after training, the solution for $\mathbf{M}$ will be part of the orthogonal matrix $\mathbf{U}$ (see \S\ref{sec:orthog}). When $\mathcal{L}_\mathit{MSP}$ is small, the transposition of $\mathbf{M}$ becomes the inverse of $\mathbf{M}$.

\subsection{Applying the Matrix Subspace Projection Framework to an AE} \label{sec:totalLoss}

To apply our matrix subspace projection (MSP) framework to an existing AE, one only needs to derive a final loss function by combining the loss of the AE and the loss of our MSP framework. 
\begin{equation} \label{eq:totalLoss}
    \mathcal{L} = \mathcal{L}_\mathit{AE} + \alpha \mathcal{L}_\mathit{MSP}
\end{equation}
where $\alpha$ is the weight for $\mathcal{L}_\mathit{MSP}$. As illustrated in Fig.~\ref{fig:model} (a) and (b), one should note that applying our framework will not change the structure of the AE, where our MSP component simply takes the latent vector $\mathbf{\hat{z}}$ of the AE as input. $\mathcal{L}_\mathit{AE}$ hopes that $\mathbf{\hat{s}}$ can store more information to reconstruct $\mathbf{x}$, but $\mathcal{L}_\mathit{MSP}$ wants $\mathbf{\hat{s}}$ to contain less information. When $\alpha$ is small, the model becomes a standard AE. When $\alpha$ is too large, the non-attribute information in $\mathbf{\hat{z}}$ is reduced excessively, resulting in the generated products tending to the average of the training samples. Therefore, another challenge we face is how to set $\alpha$ appropriately.

We propose a principled strategy for effectively determining the value of $\alpha$ (this strategy is used in all experiments in this paper). Since $\mathcal{L}_\mathit{AE}$ and $\mathcal{L}_\mathit{MSP}$ essentially represent a competing relationship for $\mathbf{\hat{z}}$ resources, we specify that $\mathcal{L}_\mathit{AE}$ and $\mathcal{L}_\mathit{MSP}$ have the same influence on updating $\mathbf{\hat{s}}$. We use $\alpha$ to represent the ``intensity'' with which the AE updates $\mathbf{\hat{z}}$ during each back-propagation process. This intensity depends on the structure of the model and the loss function used by the model.
For example, suppose an AE (for picture generation) uses a CNN decoder and L2-loss. During the training process, the error of each pixel between the generated picture and the true picture is backpropagated to $\mathbf{\hat{z}}$ as the gradients of $\mathbf{\hat{z}}$. The sum of these gradients is the final gradient of $\mathbf{\hat{z}}$ (i.e., corresponding to $\mathcal{L}_\mathit{AE}$). For a picture with $h\times w$ pixels and $c$ colour channels, there are $h\times w\times c$ parts of gradients accumulated, so the intensity is $h\times w\times c$. The intensity of $\mathbf{\hat{y}}$ for updating $\mathbf{\hat{z}}$ is the total amount of attributes (i.e. the dimension of $\mathbf{\hat{y}}$), because the error for each attribute is propagated back to $\mathbf{\hat{z}}$ and accumulated (i.e., corresponding to $\mathcal{L}_\mathit{MSP}$). Therefore, in order to balance the influence of $\mathcal{L}_\mathit{AE}$ and $\mathcal{L}_\mathit{MSP}$ on updating $\mathbf{\hat{z}}$ during training, we have: 
\begin{equation}
\alpha \approx \frac{h \times w \times c}{\text{size}(attribute) + \text{size}(\mathbf{\hat{z}})}    
\end{equation}

When using the cross-entropy-loss (or NLL loss etc.), which is usually for textual generative models (e.g. the seq2seq model), each generated word produces only one intensity, regardless of the word embedding size. 
Meanwhile, loss values returned by the cross-entropy-loss are proportional to the error, but the loss values returned by the MSP loss (which is a MSE loss) are proportional to the error's square.
Therefore, for a sentence of length $k$, the intensity of the entire sentence to $\mathbf{\hat{z}}$ is $k^2$, so that for cross-entropy-loss,
\begin{equation}
\alpha \approx \frac{k^2}{\text{size}(attribute) + \text{size}(\mathbf{\hat{z}})}  
\end{equation}

\subsection{Content Replacement and Conditional Generation}
After MSP is trained (i.e., $\mathbf{M}$ is estimated), there are two ways to perform content replacement or change of attributes. One way is to derive the orthogonal matrix $\mathbf{U} = [\mathbf{M};\mathbf{N}]$ by solving the null space  $\mathbf{N}$ of $\mathbf{M}$ (i.e., the null space is constituted of all the specific solutions for $\mathbf{n}$ w.r.t. equation $\mathbf{M}\cdot\mathbf{n}=0$, where $\mathbf{n}$ is an independent variable). Given an input $\mathbf{x}$, we first encode it as $\mathbf{\hat{z}}$. We then use $\mathbf{U}$ to obtain the attribute vector $\mathbf{\hat{y}}$ of $\mathbf{x}$ and the non-attribute information vector $\mathbf{\hat{s}}$ as follows.
\begin{equation}\label{eq:-enco}
    [\mathbf{\hat{y}};\mathbf{\hat{s}}]=[\mathbf{M};\mathbf{N}]\cdot \mathbf{\hat{z}}= \mathbf{U}\cdot \mathbf{\hat{z}}=\mathbf{U}\cdot \text{encoder}(\mathbf{x})
\end{equation}
At this point, we can directly replace $[\mathbf{\hat{y}};\mathbf{\hat{s}}]$ with $[\mathbf{y_n};\mathbf{\hat{s}}]$, where $\mathbf{y_n}$ is the new attribute vector. With $[\mathbf{y_n};\mathbf{\hat{s}}]$ and $\mathbf{U}^T$ (note that $\mathbf{U}^T$ approaches  $\mathbf{U}^{-1}$ during training), we can derive the new latent code $\mathbf{z_n}$ and then decode it into $\mathbf{x_n}$, which ultimately captures the desired new attributes. 
\begin{equation}\label{eq:-deco}
   \mathbf{x_n} = \text{decoder}(\mathbf{z_n})=\text{decoder}(\mathbf{U}^T\cdot [\mathbf{y_n};\mathbf{\hat{s}}])
\end{equation}

Alternatively, we can avoid explicitly calculating  matrix $\mathbf{N}$ (i.e. avoid calculating $\mathbf{\hat{s}}$), for content replacement. According to Eqs.\ref{eq:-enco} and \ref{eq:-deco}, we define $\mathbf{d}$ as the distance between $\mathbf{\hat{z}}$ and $\mathbf{z_n}$.
\begin{align}\label{eq:-dist}
   \mathbf{d} & = \mathbf{\hat{z}} - \mathbf{z_n} = \mathbf{U}^T\cdot [\mathbf{\hat{y}};\mathbf{\hat{s}}] - \mathbf{U}^T\cdot [\mathbf{y_n};\mathbf{\hat{s}}] \nonumber\\
   & = \mathbf{U}^T\cdot ([\mathbf{\hat{y}};\mathbf{\hat{s}}] - [\mathbf{y_n};\mathbf{\hat{s}}]) = \mathbf{U}^T\cdot [\mathbf{\hat{y}}-\mathbf{y_n};\mathbf{0}] \nonumber\\
   & = [\mathbf{M};\mathbf{N}]^T\cdot [\mathbf{\hat{y}}-\mathbf{y_n};\mathbf{0}] \nonumber\\ & = \mathbf{M}^T\cdot (\mathbf{\hat{y}}-\mathbf{y_n}) = \mathbf{M}^T\cdot (\mathbf{M}\cdot\mathbf{\hat{z}}-\mathbf{y_n})
\end{align}
It should be noted that here $\mathbf{\hat{z}}\neq \mathbf{M}^T\cdot\mathbf{M}\cdot\mathbf{\hat{z}}$ because the reconstruction loss does not allows $\mathbf{\hat{s}}$ to be zero. Thus, we have:
\begin{align}
    \mathbf{z_n} &= \mathbf{\hat{z}}-\mathbf{d} = \mathbf{\hat{z}} - \mathbf{M}^T\cdot (\mathbf{M}\cdot\mathbf{\hat{z}}-\mathbf{y_n}) \label{eq:-zn2} \\
    \mathbf{x_n} &= \text{decoder}(\mathbf{\hat{z}}-\mathbf{M}^T\cdot (\mathbf{M}\cdot\mathbf{\hat{z}}-\mathbf{y_n})) \label{eq:-xn2}
\end{align}

If the AE itself is a generative model (such as VAE), then the AE+MSP structure becomes a conditional generative model. Given a randomly sampled $\mathbf{s_r}$ and an attribute vector $\mathbf{y_r}$, the model can  generate new sample $\mathbf{x_r}$ with the desired attributes with Eq.\ref{eq:-deco}.


\begin{figure*}[!ht]
    \centering

    \includegraphics[width=17cm]{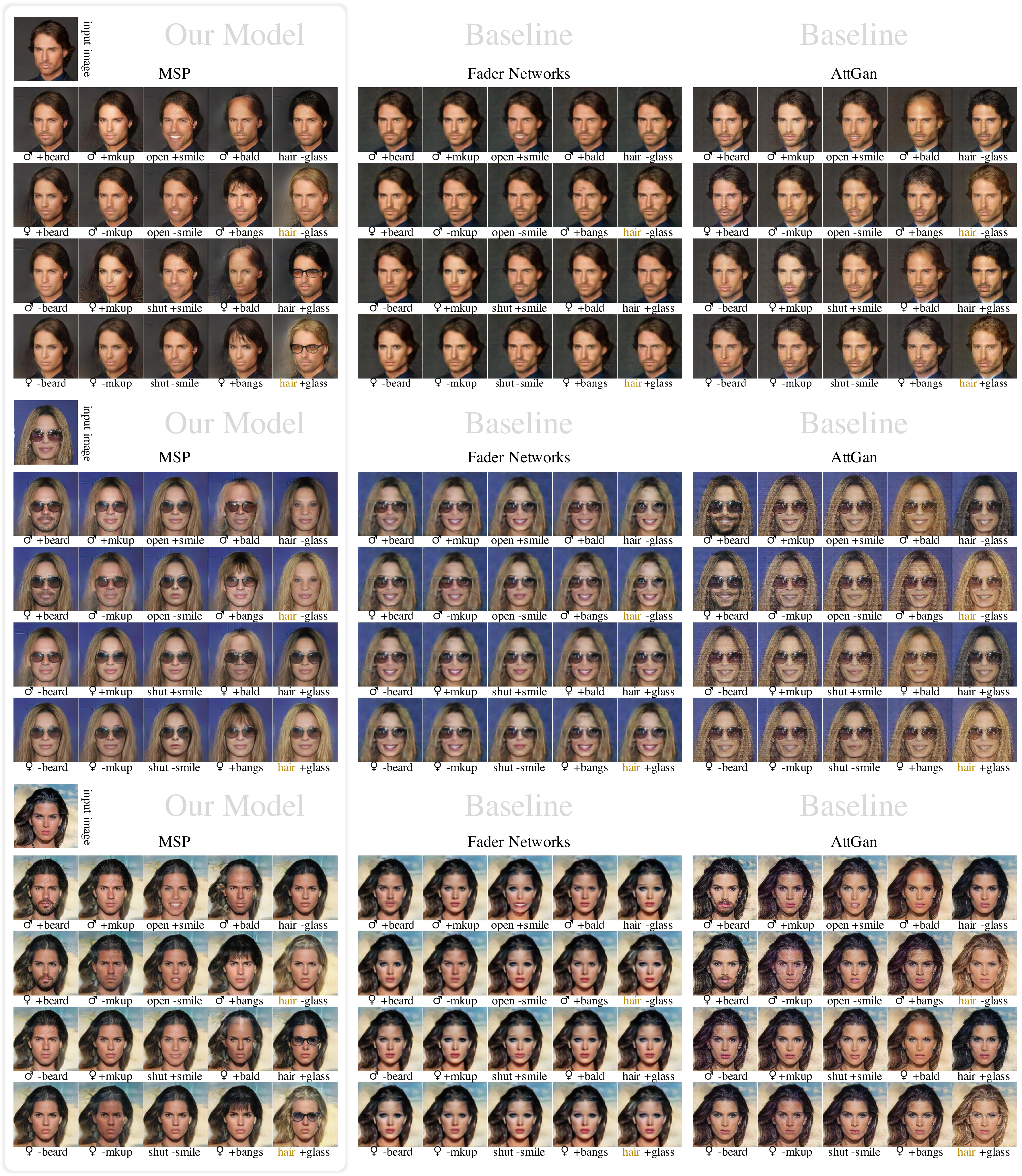}\vspace{-2ex}
    \caption{Examples of image attributes transformations using MSP (our model), Fader Networks and AttGAN.}
    \label{fig:pic_g_1}
\end{figure*}

\section{Evaluation}\label{eval}

Here we evaluate  the ability to disentangle. 
We  also evaluate the orthogonality of  $\mathbf{M}$ as it is an important indicator  of how well our algorithm can approximate $\mathbf{M}$.

\subsection{Matrix Subspace Projection in VAE}\label{se:model_vae}

We apply our model on a \textit{vanilla VAE} \cite{kingma2013auto} with the standard CNN encoder and decoder (the architectures are same as \citet{NIPS2017_7178}).
We used the ADAM optimiser with learning rate = 0.0002, mini-batch size of 256, and images are upsampled to 256 $\times$ 256.
We add an additional  PatchGAN \cite{li2016precomputed} to make the produced images sharp. The architecture of the PatchGAN discriminator also adopts the version of \citet{NIPS2017_7178}. 
Our baselines are Fader networks \cite{NIPS2017_7178} and
AttGAN  \cite{attgan}, based on their published code and settings.
We did not compare StarGAN \cite{StarGAN} because we feel it is superseded by AttGAN, which demonstrated better performance.
We did not compare RelGAN \cite{relGAN} as it does not disentangle attributes (see Fig.~\ref{relgan} (left), also RelGAN cannot add a moustache or beard to a female face, instead it will transform it to a male face with beard).

\begin{table}[tb]
\begin{center}
\begin{tabular}{|c|c|c|c|}
\hline
&	Seq2seq	&	VAE	&	VAE+GAN	\\ \hline
Better quality &	34.5\%	&	12.6\%	&	17.9\%\\
with AE only&&&		\\ \hline
Better quality	&	37.6\%	&	12.0\%	&	15.2\%	\\ 
with  AE+MSP&&&		\\ \hline
both similar quality	&	27.9\%	&	75.4\%	&	66.9\%	\\ \hline
\end{tabular}
\end{center}
\caption{Evaluation results of generation quality. Numbers in the table denote percentage of participants under the column heading who felt images were better with or without MSP. 
}
\small
\label{tab:evalu1}
\end{table}

\begin{table}[tb]
  \centering \small
  \begin{tabular}{lccc}
    \Xhline{2.5\arrayrulewidth}
    & MSP(ours) & Fader & AttGAN  \\
    \hline
    male x beard	&\textbf{0.78}	&0.42	&0.45\\
female x beard	&\textbf{0.52}	&0.03	&0.41\\
male x no-beard	&\textbf{0.86}&	0.40	&0.42\\
female x no-beard&	\textbf{0.90}	&0.61&	0.63\\
male x makeup	&\textbf{0.52}	&0.02	&0.35\\
male x no-makeup	&\textbf{0.89}	&0.50	&0.47\\
female x makeup	&\textbf{0.87}	&0.63	&0.52\\
female x no-makeup	&\textbf{0.67}	&0.42	&0.47\\
smile x open-mouth	&\textbf{0.89}	&0.59	&0.63\\
no-smile x open-mouth	&\textbf{0.66}	&0.11	&0.29\\
smile x calm-mouth	&\textbf{0.95}	&0.34	&0.33\\
no-smile x calm-mouth	&\textbf{0.76}	&0.43	&0.38\\
male x bald	&\textbf{0.78}	&0.10	&0.29\\
male x bangs	&\textbf{0.56}	&0.05	&0.19\\
female x bald	&\textbf{0.29}	&0.01	&0.17\\
female x bangs	&\textbf{0.68}	&0.21	&0.20\\
no-glasses x black-hair	&\textbf{0.74}	&0.38	&0.53\\
no-glasses x golden-hair	&\textbf{0.86}	&0.36	&0.79\\
glasses x black-hair	&\textbf{0.82}	&0.21	&0.32\\
glasses x golden-hair	&\textbf{0.77}	&0.19	&0.33\\
    \Xhline{2.5\arrayrulewidth}
  \end{tabular}
  \caption{The classification accuracy (ResNet-CNN classifier) of generated images using MSP, Fader Networks and AttGAN.}
  \label{T:cla_acc}
\end{table}
\begin{table}[htb]
\resizebox{\columnwidth}{!}{%
  \centering 
  \begin{tabular}{p{1.8cm}>{\raggedright}p{2.1cm}ccc}
    \Xhline{2.5\arrayrulewidth}
    Target  & Influence on  &	MSP &	Fader&	AttGAN\\
     attribute changed &  other attributes  & (ours)&	&	\\
    \hline
gender&	beard&	0.01&	0.28&	0.09\\
beard	&gender	&0.07	&0.11	&0.02\\
gender	&makeup	&0.02	&0.07	&0.05\\
makeup	&gender	&0.05	&0.09	&0.14\\
smile	&mouth-open&	0.01&	0.20&	0.07\\
mouth-open	&smile	&0.02	&0.07	&0.09\\

    \Xhline{2.5\arrayrulewidth}
  \end{tabular}
  }
  \caption{Quantitative evaluation  of disentanglement (using  ResNet-CNN classifier).}
  \label{T:entan}
\end{table}
\begin{table}[tb]
\begin{center}
\begin{tabular}{|c|p{1.3cm}|p{1.2cm}|p{1.7cm}|}
\hline
\multicolumn{4}{|c|}{mouth open / smiling attributes morphing}  \\ \hline
	&	Fader Network 	&	AttGAN	&		VAE+GAN MSP	(ours)\\ \hline
perfect	&	36.7\%&	47.5\%	&		68.3\%	\\ \hline
recognizable	&	20.8\%&	15.3\%	&	4.9\%	\\ \hline
unreco/unchang	&	42.5\%&	37.2\%	&	26.8\%	\\ \hline \hline
\multicolumn{4}{|c|}{male / beard attributes morphing}  \\ \hline
	&	Fader Network	&	AttGAN	&	VAE+GAN MSP	(ours)\\ \hline
perfect	&	38.3\%	&	55.9\%	&		74.4\%	\\ \hline
recognizable	&	8.3\%&	11.2\%	&		11.6\%	\\ \hline
unreco/unchang	&	53.3\%& 32.9\%	&	14.0\%	\\ \hline
\end{tabular}
\end{center}
\caption{Manual evaluation results of disentanglement. Numbers in the table denote percentage of participants under the column heading who felt the images represented the specified attribute (e.g. smiling) in a way that was perfect, recognisable, or unrecognisable/unchanged.}
\small
\label{tab:evalu2}
\end{table}

We evaluated on the CelebA dataset \cite{liu2015deep} (202,600 images) and trained one model on all 40 labelled attributes. 
The generated examples are shown in Fig.~\ref{fig:pic_g_1}.
Qualitatively  we see clear examples of what Fader networks and AttGan cannot do: For the woman with glasses (middle) FaderNetwork and AttGan show complete inability to remove the glasses; FaderNetwork completely fails to add glasses to the other two faces, and AttGan can only manage weak rims on the final woman (bottom). FaderNetwork in general struggles to change attributes, especially for the two women, while AttGan does better, but struggles with certain attributes, e.g. mostly it fails to change the final woman to male, and struggles to remove makeup.

For a quantitative evaluation we trained a classifier (ResNet-CNN) to measure the accuracy with which attributes are changed.
Table~\ref{T:cla_acc} shows that our MSP approach outperforms the competitors.
Finally we calculated the average Fr\'echet Inception Distance
(FID) \cite{frechet}
for each method: MSP=35.0, Fader=26.3, AttGAN=7.3 (lower is better, 0 is the best).
The FID score tries to calculate the similarity of original images and generated images. 
Clearly AttGAN is a winner for quality while our MSP is a winner for accuracy of attribute modification.
When AttGAN  cannot handle the attribute modification it  generates  unchanged images and can get  lower FID scores.

The results of attribute manipulation (both qualitative and quantitative) are surprisingly bad for Fader networks and AttGAN, especially relative to the examples displayed in their original papers. The primary reason for this is that we trained those models on all 40 attributes together. Fader networks works best when trained on a single attribute, as noted in Sec.~\ref{sec:relatedwork}.  The original AttGAN paper trained on 13 attributes, and indeed it performs better at attribute manipulation than Fader in our pictures. For the 40-attribute-together version, when any attribute is changed all others must be unchanged. For example, when we transition from male to female (Fig.~\ref{fig:pic_g_1} left), it is implicit that the female should keep no make-up or lipstick, etc. In the direction from female to male the male should keep no bushy eyebrows or 5 o'clock shadow. The original Fader network paper displays a beautiful example of interpolating between male and female, but the female does have make-up and lipstick and the male does have bushy eyebrows and 5 o'clock shadow.
Our difficult 40-attribute setting is central to our aims, as stated in our introduction:
we want to fully disentangle multiple attributes, because this gives a
generative model the ability to `imagine' novel examples that combine attributes in ways not present in training data.

\subsection{Human Evaluation of Generated Example Quality}

We evaluated whether our MSP model reduces the quality of  generated examples, using human evaluation via Amazon Mechanical Turk (hiring 150 participants in total).

For each model, we randomly generated 1,000 example pairs. Each pair contains a reconstructed example (from AE) and an example generated by AE+MSP with one or two random attribute modification (attributes were changed to $-1$ if they were originally $> 0$, or changed to 1 if they were $< 0$)\footnote{The generated examples were automatically filtered to prevent conflict attributes; e.g. images of woman with beard are not provided to the participants in this experiment.}. The participants were shown the examples pair-by-pair in the blind test, and they were asked to \textit{please choose the one with better text/image quality, or choose both if you think they perform similarly}. The participants were told that the text quality means the fluency, semantic accuracy, and syntactic accuracy, and the image quality means the clarity and (face) recognisability. The results are shown in Table~\ref{tab:evalu1}.
We treat the scores (i.e. participants' choices) of the result as a Likert scale, and we set our null hypothesis to be $H_0$: \textit{generation quality of AE+MSP is worse than using the AE only}, and, $H_1$: \textit{generation quality of AE+MSP is equal or higher than using the AE only}. The hypotheses are tested by a discrete Mann-Whitney U-test, rejecting $H_0$  with $p<0.03$.

\subsection{Evaluation of Disentanglement}

Disentanglement is also an important feature of our model. It means that when an attribute is modified, other attributes remain unchanged. 
We make the three models (VAE+GAN+MSP, AttGAN, and Fader Networks) generate images by manipulating two groups of highly correlated attributes, openness of mouth / smiling, and male / beard. 
For the two groups, the three models should respectively generate the images with closed mouth $\times$ no smiling, closed mouth $\times$ smiling, opened mouth $\times$ no smiling, opened mouth $\times$ smiling, female $\times$ no beard, female $\times$ beard, male $\times$ no beard, and male $\times$ beard. We hired 50 participants in Amazon Mechanical Turk; each of them was given 40 image blocks. A block contains four images, which were from the mouth / smiling group or the male / beard group, and which were generated by one of the three models. The participants were told which image should represent which attributes, and the participants evaluated whether it did for  each image in the block by using a 3-level Likert scale (perfect, recognisable, and unrecognisable/unchanged). The results are shown in Table \ref{tab:evalu2}. It shows that our model performs significantly better than the baseline. ($p<0.0001$).

\begin{table*}[tb]
\begin{center}
\begin{tabular}{|l|p{6.5cm}|>{\raggedright\arraybackslash}p{7.4cm}|}
\hline
& Example 1 & Example 2
\\ \hline
Orig-attribute 
& eatType[pub], customer-rating[5-out-of-5], name[Blue-Spice], near[Crowne-Plaza-Hotel]
& familyFriendly[yes], area[city-centre], eatType[pub], food[Japanese], near[Express-by-Holiday-Inn], name[Green-Man]
\\ \hline
Orig-text 
&  the blue spice pub , near crowne plaza hotel , has a customer rating of 5 out of 5 .
& near the express by holiday inn in the city centre is green man . it is a japanese pub that is family-friendly .
\\ \hline
New-attribute 
& eatType[coffee-shop], customer-rating[5-out-of-5], name[Blue-Spice], near[Avalon]
& familyFriendly[no], area[riverside], eatType[coffee-shop], food[French], near[The-Six-Bells], name[Green-Man]
\\ \hline
New-text
& the blue spice coffee shop , near avalon has a customer rating of 5 out of 5 .
& near the six bells in the riverside area is a green man . it is a french coffee shop that is not family-friendly .
\\ \hline
\end{tabular}
\end{center}
\caption{Results of changing attributes in E2E corpus.}
\small
\label{tab:exp-text}
\end{table*}

\begin{figure*}[tb]
    \centering
    \includegraphics[width=16.5cm]{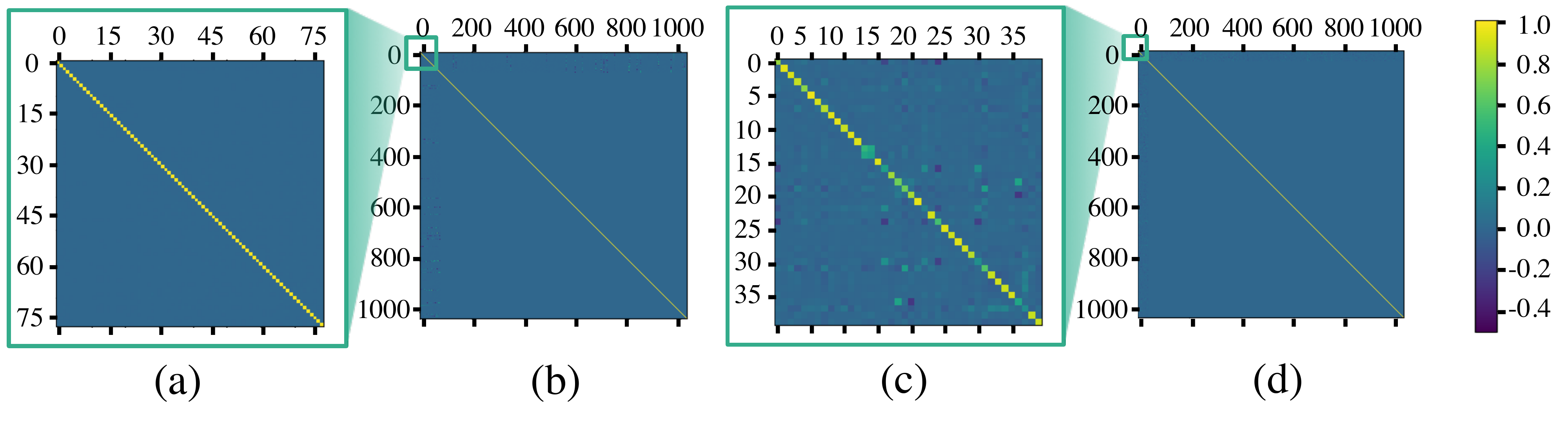}
    \caption{Measuring orthogonality: Heat map of $\mathbf{M}^{T}\cdot \mathbf{M}$ and $\mathbf{U}^{T}\cdot \mathbf{U}$ for Seq2seq+MSP (a,b), and VAE+GAN+MSP (c,d)}
    \label{fig:orthog}
\end{figure*}

In addition to human evaluation, we also conducted a quantitative evaluation to test how well a model can change an attribute in isolation. For some selected  highly correlated attributes we change one target attribute, and measure the change in another non-target attribute. For instance (the row of gender/beard in Table~\ref{T:entan}), when the gender attribute is changed manually, we measure the amount by which the beard attribute is consequently changed. The results are shown in Table~\ref{T:entan}. 
Note that, the scores show how much the non-target attributes are affected, but not whether the target attributes are correctly changed in the generated pictures. Therefore the scores need to  be read in conjunction with Table~\ref{T:cla_acc}. According to both Table~\ref{T:cla_acc} and Table~\ref{T:entan}, we can conclude that in both of the aspects of the manipulation of attributes and avoiding influence on non-target attributes, the performance of our model exceeds the baselines.

\subsection{Matrix Subspace Projection in Seq2seq}

We apply our model to a classic \textit{seq2seq} model for textual content replacement, in order to determine if we can replace words according to the given attributes and keep other words unchanged.
In this task, we adopt the E2E corpus \cite{dusek2019e2e}, which contains 50k+ reviews of restaurants (E2E dataset is developed for Natural Language Generation, but here we use it for content replacement). Each review is a single sentence that is labelled by the attribute-value pairs, for example, ``name=[The Eagle]'', ``food=[French]'', and ``customerRating=[3/5]''. We regard each attribute-value pair as a unique label. All the attributes constitute $y$ whose entries are 1 or 0 to represent each value  (the correct texts of the attribute name or value are NOT used).

Both the encoder and decoder of the seq2seq model are formed by two-layer LSTMs. The model is trained for 1000 epochs (on a Tesla T4 around 12 hours). After training, we reconstruct the review sentences with randomly replaced attributes, for example replacing ``name=[The Eagle]'' by ``name=[Burger King]'', ``customerRating=[3/5]'' by ``customerRating=[1/5]''. 50\% of attributes are changed in each sentence.
The outcomes are shown in Table \ref{tab:exp-text}.

\subsection{Orthogonality Evaluation} \label{sec:orthog}

The ability to disentangle attributes is ensured by the orthogonality of $\mathbf{M}$ in our model. Instead of directly using an orthogonal matrix, we train $\mathbf{M}$ to be orthogonal. Thus, we evaluate how close $\mathbf{M}$ is to the orthogonal matrix.
Fig.~\ref{fig:orthog} shows the heat map of $\mathbf{M}^{T}\cdot \mathbf{M}$ and $\mathbf{U}^{T}\cdot \mathbf{U}$, which indicates that the production is fairly close to a unit matrix. It visualises $\mathbf{M}^T\cdot \mathbf{M}$ in the seq2seq version of our model (Fig.~\ref{fig:orthog} (a)) and in the VAE version (Fig.~\ref{fig:orthog} (c)). The matrix $\mathbf{U}^{T}\cdot \mathbf{U}$ ($\mathbf{U}$ is formed by $\mathbf{M}$ concatenating its null space) is also visualised (in Fig.~\ref{fig:orthog} (b) and (d)). It is clear that when the matrices are multiplied by their transposes, the products do approximate the unit matrix.
Although Fig.~\ref{fig:orthog} (c) shows that a small number of attributes remain slightly entangled (by the green and deep purple pixels), this is mainly caused by the few conflicting attributes in CelebA, for example the receding-hairline~$\times$~bald~$\times$~bangs, and the straight-hair~$\times$~wavy-hair.
Thus, $\mathbf{M}$ is indeed trained to be a (partial) orthogonal matrix.

\section{Conclusion}
We propose a matrix projection plugin that can be attached to various autoencoders (e.g. Seq2seq, VAE) to make the latent space factorised and  disentangled, based on  labelled attribute information, which ensures that manipulation in the latent space is much easier. 
We test the attribute manipulation ability of our model on an image dataset and text corpus, obtaining results that show clean disentanglement. 
In addition our model involves a simpler training process than adversarial approaches which need a long training with a very low weight on the loss coming from the discriminator that removes attribute information, to avoid the encoder being too affected by this loss term \cite{NIPS2017_7178}.

\section*{Acknowledgement}
We would like to thank all the anonymous reviewers for their insightful comments. This work is supported by the award made by the
UK Engineering and Physical Sciences Research
Council (Grant number: EP/P011829/1).

\bibliography{egbib}
\bibliographystyle{icml2020}

\end{document}